\documentclass[12pt,authoryear]{./elsarticle}
\usepackage{graphicx} 
\usepackage{url}
\usepackage{subfigure}
\usepackage{makecell}
\usepackage{threeparttable}
\usepackage{multirow}
\usepackage{amsthm}
\usepackage{mathrsfs}
\usepackage{tabularray}
\usepackage{bbding}
\usepackage{arydshln}
\usepackage{amssymb}
\DeclareMathAlphabet{\mathcal}{OMS}{cmsy}{m}{n}
\usepackage{pifont}% http://ctan.org/pkg/pifont
\newcommand{\cmark}{\ding{51}}%
\newcommand{\xmark}{\ding{55}}%
\usepackage{float}
\usepackage{array}
\usepackage{booktabs}
\usepackage{soul}
\usepackage{color}
\usepackage[caption=false,font=normalsize,labelfont=sf,textfont=sf]{subfig}
\usepackage{tabularx}
\newcolumntype{Y}{>{\centering\arraybackslash}X}
\usepackage{amsmath}
\usepackage{amssymb}
\usepackage{bm}
\usepackage{adjustbox}

\begin{document}

\begin{frontmatter}

%% Title, authors and addresses

%% use the tnoteref command within \title for footnotes;
%% use the tnotetext command for theassociated footnote;
%% use the fnref command within \author or \address for footnotes;
%% use the fntext command for theassociated footnote;
%% use the corref command within \author for corresponding author footnotes;
%% use the cortext command for theassociated footnote;
%% use the ead command for the email address,
%% and the form \ead[url] for the home page:
%% \title{Title\tnoteref{label1}}
%% \tnotetext[label1]{}
%% \author{Name\corref{cor1}\fnref{label2}}
%% \ead{email address}
%% \ead[url]{home page}
%% \fntext[label2]{}
%% \cortext[cor1]{}
%% \affiliation{organization={},
%%             addressline={},
%%             city={},
%%             postcode={},
%%             state={},
%%             country={}}
%% \fntext[label3]{}

\title{Enabling Generalized Zero-shot Learning Towards Unseen Domains by Intrinsic Learning from Redundant LLM Semantics}

%% use optional labels to link authors explicitly to addresses:
%% \author[label1,label2]{}
%% \affiliation[label1]{organization={},
%%             addressline={},
%%             city={},
%%             postcode={},
%%             state={},
%%             country={}}
%%
%% \affiliation[label2]{organization={},
%%             addressline={},
%%             city={},
%%             postcode={},
%%             state={},
%%             country={}}

% \author{Jiaqi Yue, Chunhui Zhao*, Jiancheng Zhao, Biao Huang}

% \cortext[1]{Corresponding author. E-mail address: chhzhao@zju.edu.cn} 
% \address{College of Control Science and Engineering, Zhejiang University, \\Hangzhou 310027, China}

\author[a]{Jiaqi Yue}
\author[a]{Chunhui Zhao\corref{cor1}}
\author[a]{Jiancheng Zhao}
\author[b]{Biao Huang}

\address[a]{College of Control Science and Engineering, Zhejiang University, \\Hangzhou 310027, China}
\address[b]{Department of Chemical and Materials Engineering, University of Alberta, \\Edmonton, AB T6G 1H9, Canada}

\cortext[cor1]{Corresponding author. E-mail address: chhzhao@zju.edu.cn}

\begin{abstract}
  Generalized zero-shot learning (GZSL) focuses on recognizing seen and unseen classes against domain shift problem where data of unseen classes may be misclassified as seen classes. However, existing GZSL is still limited to seen domains. In the current work, we study cross-domain GZSL (CDGZSL) which addresses GZSL towards unseen domains. Different from existing GZSL methods, CDGZSL constructs a common feature space across domains and acquires the corresponding intrinsic semantics shared among domains to transfer from seen to unseen domains. Considering the information asymmetry problem caused by redundant class semantics annotated with large language models (LLMs), we present Meta Domain Alignment Semantic Refinement (MDASR). Technically, MDASR consists of two parts: Inter-class similarity alignment, which eliminates the non-intrinsic semantics not shared across all domains under the guidance of inter-class feature relationships, and unseen-class meta generation, which preserves intrinsic semantics to maintain connectivity between seen and unseen classes by simulating feature generation. MDASR effectively aligns the redundant semantic space with the common feature space, mitigating the information asymmetry in CDGZSL. The effectiveness of MDASR is demonstrated on two datasets, Office-Home and Mini-DomainNet, and we have shared the LLM-based semantics for these datasets as a benchmark. 

\end{abstract}

%%Graphical abstract
% \begin{graphicalabstract}
% %\includegraphics{grabs}
% \end{graphicalabstract}

%%Research highlights
% \begin{highlights}
% \item Systematically evaluate ten representative embedding-aware generative models (EAGMs) for any-shot learning.
% \item Demonstrate the significant impact of simple modifications on embedding features, enhancing the performance of EAGMs for any-shot learning.
% \item Introduce the generative any-shot learning (GASL) repository, providing models, features, parameters, and settings for EAGMs in any-shot learning.
% \end{highlights}


\end{frontmatter}

%% \linenumbers

\section{Introduction}

The era of Large Language Models (LLMs) marks a tremendous leap in the ability of artificial intelligence to understand and generate natural language. With massive parameters and extensive training data, LLMs have the capacity to internalize vast amounts of knowledge and externalize it for application in the virtual and real world \citep{zhao2023survey, 10529537}. In some machine learning studies that fuse data and knowledge \citep{pan2025attribute, 10040605}, researchers previously relied on the knowledge from human experts to provide models with prior information, aiming to obtain results that outperform purely data-driven techniques \citep{lu2025multiple, zhao2022perspectives}. The emerging LLMs have, to some extent, supplanted the role of human experts, thereby bringing promising prospects to data-knowledge fusion studies, such as zero-shot learning (ZSL) \citep{8385127, gautam2022tf, song2024multi}.

ZSL is designed to recognize unseen classes not present in the training set by leveraging semantic knowledge to bridge the gap between seen and unseen classes, mimicking human cognitive inference \citep{9072621, lampert2013attribute, 10428069}. Initially, ZSL approaches like DAP \citep{lampert2013attribute} and ALE \citep{akata2015label} construct mappings from data to semantics, training on seen classes and extrapolating to unseen test data, under the assumption that only unseen classes would be tested. This works well when the assumption holds, but real-world scenarios often involve a mix of seen and unseen classes. To overcome this limitation, Generalized Zero-Shot Learning (GZSL) \citep{feng2022bias,zhang2025generalized} is introduced, requiring models to identify both seen and unseen classes during testing. However, ZSL models tend to misclassify unseen classes as seen due to the domain shift problem \citep{9371418, 10443220}, leading to poor GZSL performance. To combat this, generative methods such as ZeroDiff \citep{ye2024exploring} and VADS \citep{hou2024visual} have been developed, modeling between the semantic space and the feature space, generating unseen class data based on semantic knowledge to reduce model overfitting on seen classes and mitigate domain shift problem.

Before the era of LLMs, ZSL and GZSL acquired semantic knowledge through the method of expert-annotated attributes \citep{sun2021research, rossi2024generalizability}. This approach is only feasible for datasets with a relatively small number of classes due to the high cost of annotation. Some professional fields even require seeking knowledge from experts \citep{feng2020fault}. For large-scale datasets, researchers often resorted to using word vectors as a means of automatically acquiring knowledge of different classes \citep{sun2021research}. However, the semantic information contained within these vectors is not rich enough, leading to potentially suboptimal experimental results. Fortunately, as mentioned earlier, the arrival of LLMs has provided a more efficient way to acquire knowledge \citep{naeem2023i2mvformer}. By crafting appropriate prompts, we can extract a vast amount of descriptive text for each class from an LLM, and the rich semantic information contained in these texts allows for improving the performance of ZSL at a lower cost.

Recently, the simultaneous exploration of data from unseen domains and unseen classes has garnered considerable attentions, which is called cross-domain zero-shot learning (CDZSL) \citep{mancini2020towards}. The training set for CDZSL comprises data from seen classes across multiple seen domains, while the testing set consists of data from unseen classes within the unseen domain. Undoubtedly, because of the intervention of domains, CDZSL presents a greater challenge than ZSL due to its demand for a heightened level of model generalization. To address the gap of unseen classes and domains, Mancini et al. employed the Mixup technique \citep{zhang2018mixup} and introduced CuMix \citep{mancini2020towards}. This approach utilizes the underlying structure of ZSL methods, mapping data to the feature space before subsequently mapping features to the semantic space. During the training process, samples from diverse domains and classes are intermixed within the data and feature spaces, effectively simulating novel domains and classes. Building upon CuMix, SEIC \citep{mondal2022seic} introduced intermediate classes to add the feature embedding space with semantic meaning. 
% The semantic knowledge utilized by these methods is still in the form of word vectors.

\begin{figure}[htbp]
        \centering
        \includegraphics[width=\columnwidth]{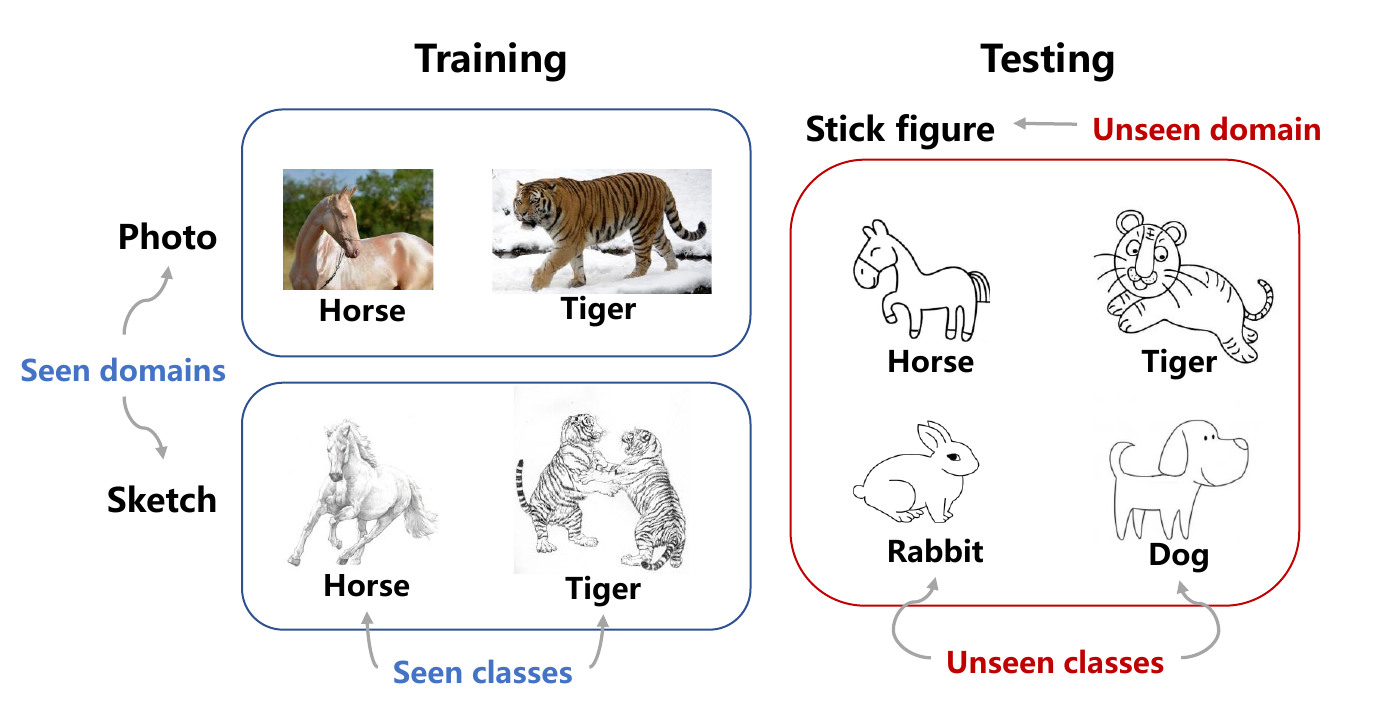}
        \caption{The setting of CDGZSL. The training set includes seen classes in seen domains, and the testing set includes both seen and unseen classes in unseen domains.}
        \label{fig: CDGZSL}
\end{figure}

Just like in ZSL, CDZSL also makes an assumption about the inference process: during the testing phase, only unseen classes from unseen domains are present. We believe that the assumption of CDZSL does not agree well with reality, as in real-world scenarios, both seen and unseen classes are likely to occur. Hence, we first introduce Cross-Domain Generalized Zero-Shot Learning (CDGZSL) and demonstrate it in Fig. \ref{fig: CDGZSL}. In CDGZSL, the training set includes seen class data from multiple seen domains, while the test set includes both seen and unseen class data from the unseen domain. A comparison between zero-shot learning, generalized zero-shot learning, cross-domain zero-shot learning, and cross-domain generalized zero-shot learning can be seen in Table \ref{Tab: comparison}. Existing studies encompass zero-shot learning, generalized zero-shot learning and cross-domain zero-shot learning. Generalized zero-shot learning removes the inference assumption of zero-shot learning but maintains the assumption of the single domain. In contrast, cross-domain zero-shot learning, despite its ability to handle multiple domains, does not eliminate the inference assumption. This demonstrates that generalized zero-shot learning and cross-domain zero-shot learning both have assumptions, and cross-domain generalized zero-shot learning breaks the limitations of both, resulting in test scenarios that agree well with reality.

\begin{table}[htbp]
        \centering
        \small
        \caption{Comparison among several ZSL settings.}
        \label{Tab: comparison}
        \begin{threeparttable}
        \begin{tabularx}{\columnwidth}{c>{\centering\arraybackslash}X>{\centering\arraybackslash}X>{\centering\arraybackslash}X>{\centering\arraybackslash}X}
        \hline
        \textbf{Setting} & \textbf{Existing Research} & \textbf{Domain} & \textbf{Inference Class} & \textbf{Assumption}   \\
        \hline
        ZSL & \cmark & Single & Unseen class only & \cmark  \\
        
        GZSL & \cmark & Single & Seen \& unseen class & \cmark \\
        
        CDZSL & \cmark & Multiple & Unseen class only & \cmark \\
        
        CDGZSL & \xmark & Multiple & Seen \& unseen class & \xmark \\
        \hline
        \end{tabularx}
        \begin{tablenotes}
                \footnotesize
                \item[*] \cmark indicates the presence of research or assumption, while \xmark represents the opposite meaning.
              \end{tablenotes}
        \end{threeparttable}
        \end{table}

Although generative methods could be employed to address the transition from zero-shot learning to generalized zero-shot learning, the same approach cannot be directly applied from cross-domain zero-shot learning to cross-domain generalized zero-shot learning. As mentioned earlier, a large volume of class description text can be obtained from LLMs as semantics for the sake of cost-effectiveness and speed, and is applied to all domains (including seen and unseen domains). However, this leads to the asymmetry between the low information density of semantic modality and the high information density of feature modality. In terms of the semantic modality, a significant portion of the large-volume LLM semantics is not intrinsic semantics which are shared across domains, leading to a low knowledge information density. For the feature modality, in order to generalize from seen domains to unseen domains, the generator needs to produce features within a common feature space shared by all domains. However, this common feature space encapsulates the essence of each class and thus has a higher information density. This asymmetry poses a challenge for the generator to accurately model the relationship between these two modalities. It results in greater differences between the generated features for unseen classes and their actual features, making it difficult to address the domain shift problem when transitioning from CDZSL to CDGZSL.

In this paper, we introduce a novel approach named Meta Domain Alignment Semantic Refinement (MDASR) aimed at enabling GZSL towards unseen domains. To address the aforementioned issue of information asymmetry, unlike what the existing ZSL, GZSL and CDZSL endeavors to, we learn intrinsic information from the redundant semantics and perform generative modeling between the refined semantics and domain common features, which are symmetrical in information density, to mitigate the domain shift problem. To this end, MDASR includes two optimization objectives: Inter-class Similarity Alignment (ISA) and Unseen-class Meta Generation (UMG). Guided by the feature space, ISA eliminates non-intrinsic semantics while UMG preserves intrinsic semantics. Specifically, on the one hand, ISA leverages the fundamental and invariant relationships between classes, which are derived from multi-domain data, as a standard to remove non-intrinsic semantics. On the other hand, UMG simulates the generation process of unseen class features to ensure that the semantic linkage from seen to unseen classes remains unbroken, thus preserving intrinsic semantics. In summary, our method, through ISA and UMG, addresses the issue of information asymmetry between semantic and feature spaces, thereby alleviating the domain shift problem in CDGZSL. The total contributions of this paper are summarized below.

\begin{enumerate}
\item  We have introduced CDGZSL firstly, overcoming the inability of CDZSL towards unseen domains to recognize seen and unseen classes. Furthermore, we have uncovered the bottleneck of CDGZSL, i.e., information asymmetry caused by semantic redundancy, and addressed it from the perspective of intrinsic semantic learning. 

\item  A novel MDASR framework is designed to eliminate non-intrinsic semantics and preserve intrinsic semantics through ISA and UMG. This method refines the low-information-density semantic space under the guidance of the high-information-density feature space.

\item We have verified the CDGZSL capability of MDASR on two open-source datasets, Office-Home and Mini-DomainNet, and we have then made the corresponding LLM-based semantic texts for these datasets available to the public, providing a benchmark for interested researchers.
\end{enumerate}

\section{Motivation}
\label{sec: motivation}
In this section, we will clarify our motivation: the problem of semantic variation and redundancy, and the resulting information asymmetry between semantic space and feature space that makes the solution of domain shift problem in CDGZSL challenging.

\subsection{The problem of semantic variation and redundancy}

We introduce the semantic variation caused by multiple domains, as well as the issue of redundancy arising from the general semantic descriptions.

\subsubsection{Semantic variation}

\begin{figure}[htbp]
        \centering
        \includegraphics[width=0.7\columnwidth]{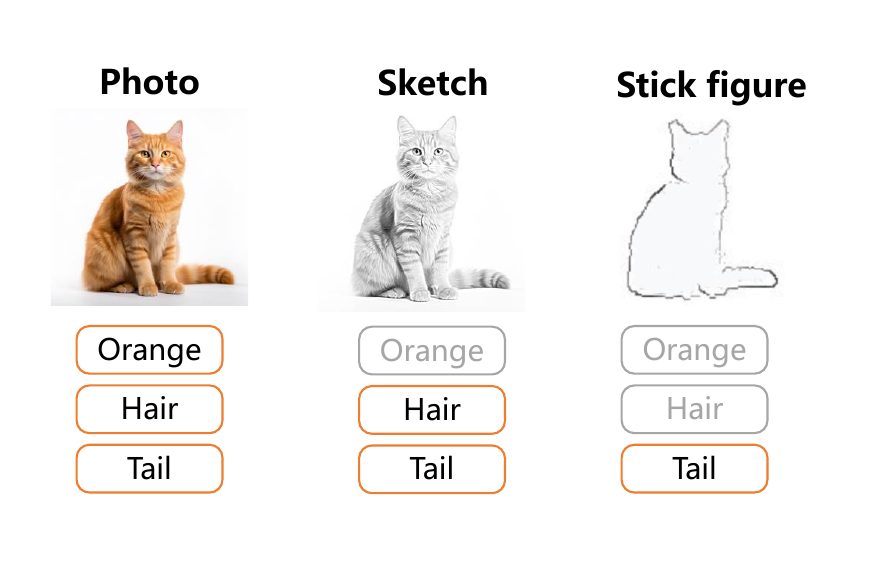}
        \caption{An example of semantic variation.}
        \label{fig: statement1}
\end{figure}

Data from multiple domains represent different manifestations of the same class, resulting in semantics that are valid in a specific domain often being invalid in different domains. We demonstrate this concept in Fig. \ref{fig: statement1} by leveraging a simple example. When describing an orange cat, the following semantics could be given: orange, with hair, and with a tale. In the domain of photos, these semantics are all valid. However, in the domain of sketches, the color-related semantic `orange' becomes invalid. Moreover, the texture-related semantic `with hair' is not applicable as well in the domain of stick figures. This indicates that the semantics associated with a class can vary as the domain changes.

\subsubsection{Semantic redundancy}

\begin{figure}[htbp]
        \centering
        \includegraphics[width=\columnwidth]{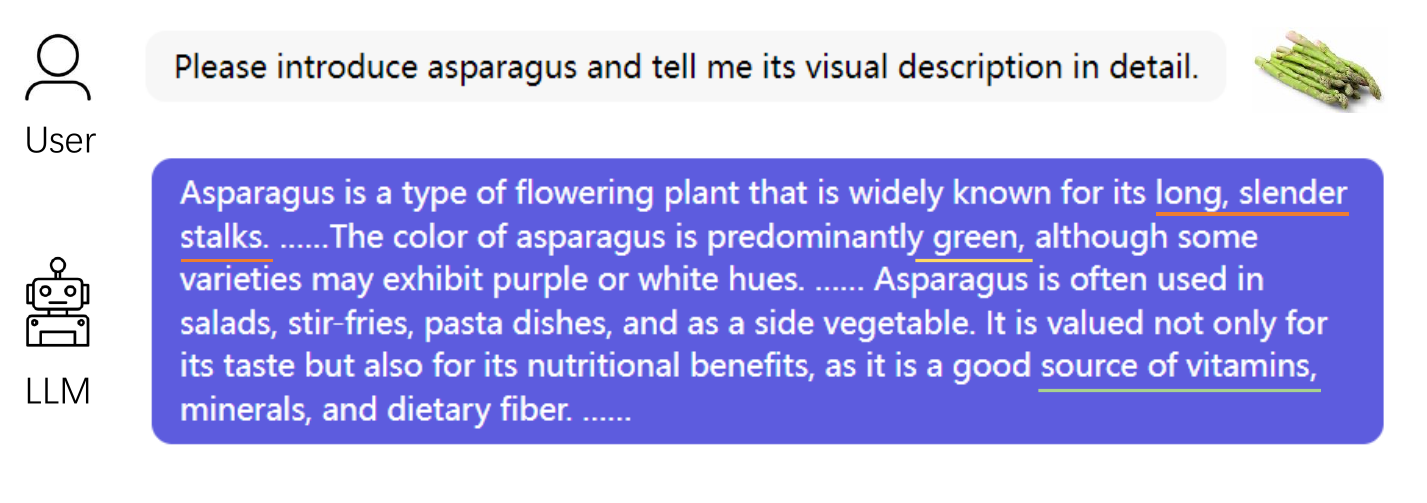}
        \caption{Demonstration of the redundancy in semantic annotations by LLMs. The orange line represents intrinsic semantics, the yellow line may disappear due to domain influence, and the green line is completely unrelated to vision.}
        \label{fig: statement1.56}
\end{figure}

To capture the semantics of a class, we could employ LLMs for their comprehensive descriptions. The semantic descriptions labeled by LLMs are usually extensive texts containing general information about the object. However, this also leads to a lot of redundant information. We provide an example in Fig. \ref{fig: statement1.56}, where we asked the LLM to provide the semantics of asparagus. The semantics provided include shape, color, and nutrition. Shape generally remains unchanged and is the intrinsic semantic; color may become redundant with changes in domain; nutrition is unrelated to visual features and is definitively redundant information.

\begin{figure}[htbp]
        \centering
        \includegraphics[width=\columnwidth]{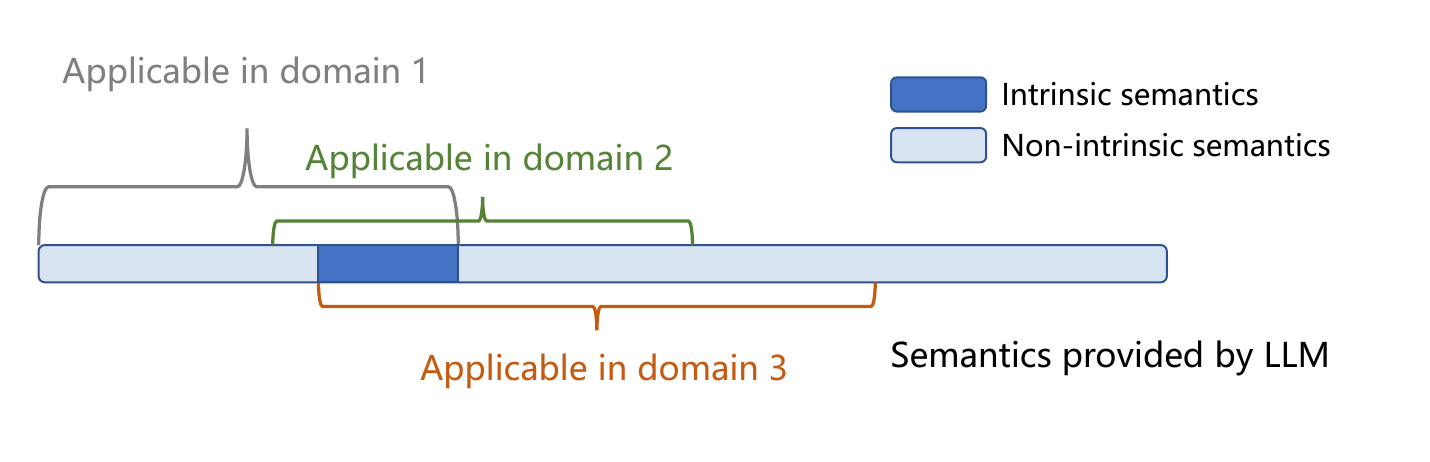}
        \caption{The redundancy of LLM semantics. Only a part of semantics is shared across domains.}
        \label{fig: statement1.5}
\end{figure}

As shown in Fig. \ref{fig: statement1.5}, for the large volume of semantics provided by LLMs (represented by the light blue bar), only a part of the semantics are intrinsic (represented by the dark blue bar) because they are the intersection of effective semantics across various domains. Therefore, although LLMs can provide comprehensive descriptions, they are less effective for cross-domain transfer because they fail to capture the intrinsic semantics.

\textbf{Remark}: In the present CDZSL and our CDGZSL, general class descriptions are obtained and applied directly to all domains, thereby resulting in the above semantic redundancy. Precise class descriptions for each domain still demand extensive expert review and thus are beyond our current scope.

\subsection{Information asymmetry between semantics and features}

\begin{figure}[htbp]
        \centering
        \includegraphics[width=\columnwidth]{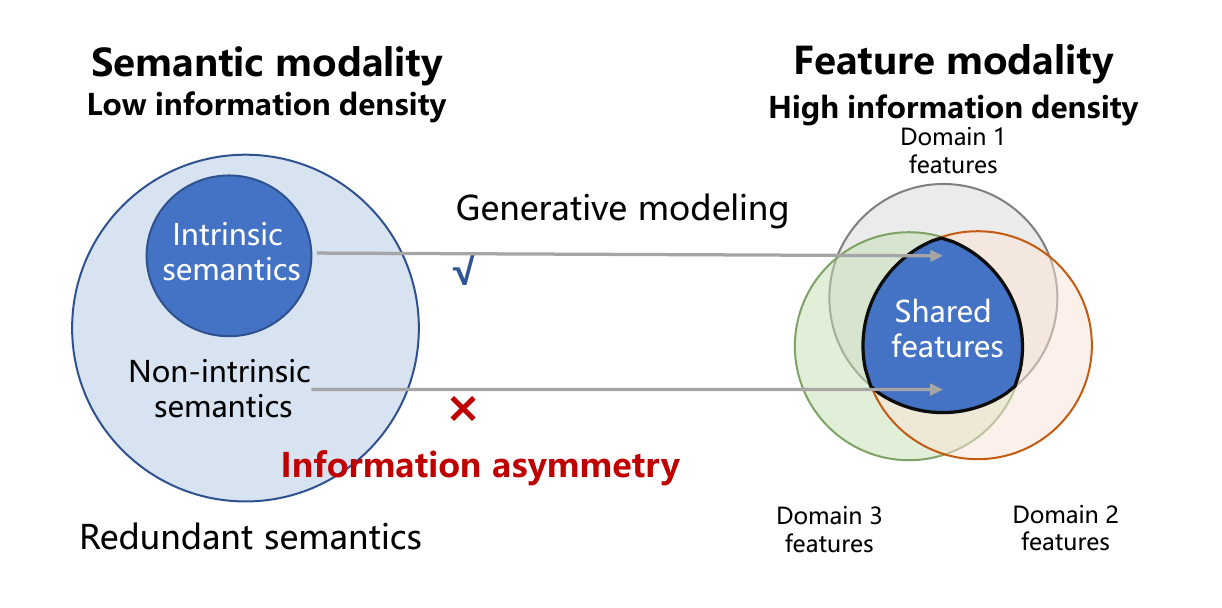}
        \caption{The information asymmetry between semantic and feature modality. The non-intrinsic semantics have no corresponding part in the shared features.}
        \label{fig: statement2}
\end{figure}

If we want to address domain shift problem, a generator is needed to model the relationship between the modalities of semantics and features, and generate features of unseen classes \citep{pourpanah2022review}. However, there is an information asymmetry between the two modalities, as shown in Fig. \ref{fig: statement2}, impeding the effectiveness of generative methods in transitioning from CDZSL to CDGZSL. 

The information asymmetry between the semantic and feature modalities originates from their different information densities. In this paper, information density is defined as the proportion of shared information across domains relative to all information. For the semantic modality, the intrinsic semantics shared by each domain only account for a portion of the general semantics provided by the LLM, while the rest of the information is non-intrinsic, resulting in low information density, as shown in the left half of Fig. \ref{fig: statement2}. In contrast, for the feature modality, domain adaptation methods can extract the common feature space across domains, eliminating domain-specific information, thus leading to high information density, as represented by the dark blue overlapping area in the right half of Fig. \ref{fig: statement2}.

% In fact, the solution lies in obtaining the common features and identifying the corresponding intrinsic information within the semantics. The prior problem has been resolved by predecessors, making it simpler, whereas the latter remains unsolved, presenting more of a challenge. If the information asymmetry is not tackled, the generator could only be trained with non-intrinsic semantics mixed together, which harms addressing the domain shift problem.

Existing approaches to enhance model transferability include transfer learning \citep{ganin2016domain,lu2025multiple} and meta-learning \citep{finn2017model,nichol2018first}. However, these methods face significant limitations in addressing the information asymmetry caused by semantic redundancy. Transfer learning focuses on feature alignment but ignores the redundancy in semantic annotations. Meta-learning optimizes for task-specific adaptation but does not consider cross-domain semantic variations. The previously mentioned zero-shot learning, while leveraging semantic knowledge, assumes that semantics are domain-agnostic, failing to handle domain-specific redundancy. As summarized in Table \ref{tab:df-compare}, none of these methods explicitly address the information asymmetry between semantic and feature modalities, which is critical for cross-domain generalized zero-shot learning (CDGZSL).

\begin{table}[h!]
        \centering
        \small
        \begin{adjustbox}{width=\textwidth}
        % \begin{tabular}{c|m{90pt}|m{120pt}|m{80pt}}
        \setcellgapes{5pt}
        \begin{tabular}{c|c|c|c}
            \hline
            \textbf{Paradigm} & \makecell[c]{\textbf{Representative}\\ \textbf{Works}} & \textbf{Problem Solved} & \makecell[c]{\textbf{Consider}\\ \textbf{Information}\\ \textbf{Asymmetry?}} \\ \hline
            Transfer Learning & 
            \makecell*[l] {$\cdot$ DANN \citep{ganin2016domain} \\ $\cdot$ MAN \citep{lu2025multiple}} & 
            \makecell{Align different \\domains} & 
            \makecell*[c] {\xmark} \\ \hline
            
            Meta-Learning & 
            \makecell*[l]{$\cdot$ MAML \citep{finn2017model} \\$\cdot$ Reptile \citep{nichol2018first}} & 
            \makecell{Fast adaptation\\ to new tasks} & 
            \makecell*[c] {\xmark} \\ \hline
            
            Existing ZSL & 
            \makecell*[l]{$\cdot$ DAP \citep{lampert2013attribute} \\ $\cdot$ CuMix \citep{mancini2020towards} \\ $\cdot$ Zerodiff \citep{ye2024exploring}} & 
            \makecell{Recognize \\unseen classes} & 
            \makecell*[c] {\xmark} \\ \hline

            \makecell*[c]{CDGZSL} & \makecell[l]{$\cdot$ MDASR\\(Proposed)} & \makecell{Refine redundant \\semantics, thus recognize\\ seen \& unseen classes\\ in useen domain} & 
            \makecell*[c] {\checkmark} \\ \hline
        \end{tabular}
\end{adjustbox}
        \caption{Comparison between existing paradigms and the proposed MDASR method. \checkmark indicates that the method consider information asymmetry, while \xmark represents the opposite.}
        \label{tab:df-compare}
    \end{table}

    As shown in Table \ref{tab:df-compare}, existing paradigms cannot be simply combined to solve the problem of information asymmetry in CDGZSL. Therefore, a novel approach is required to explicitly address this problem. Our proposed MDASR achieves this by refining redundant semantics and preserving intrinsic ones, thereby mitigating information asymmetry and enabling effective CDGZSL. We will explain our method in the following section.

\begin{figure*}[htbp]
        \centering
        \includegraphics[width=\textwidth]{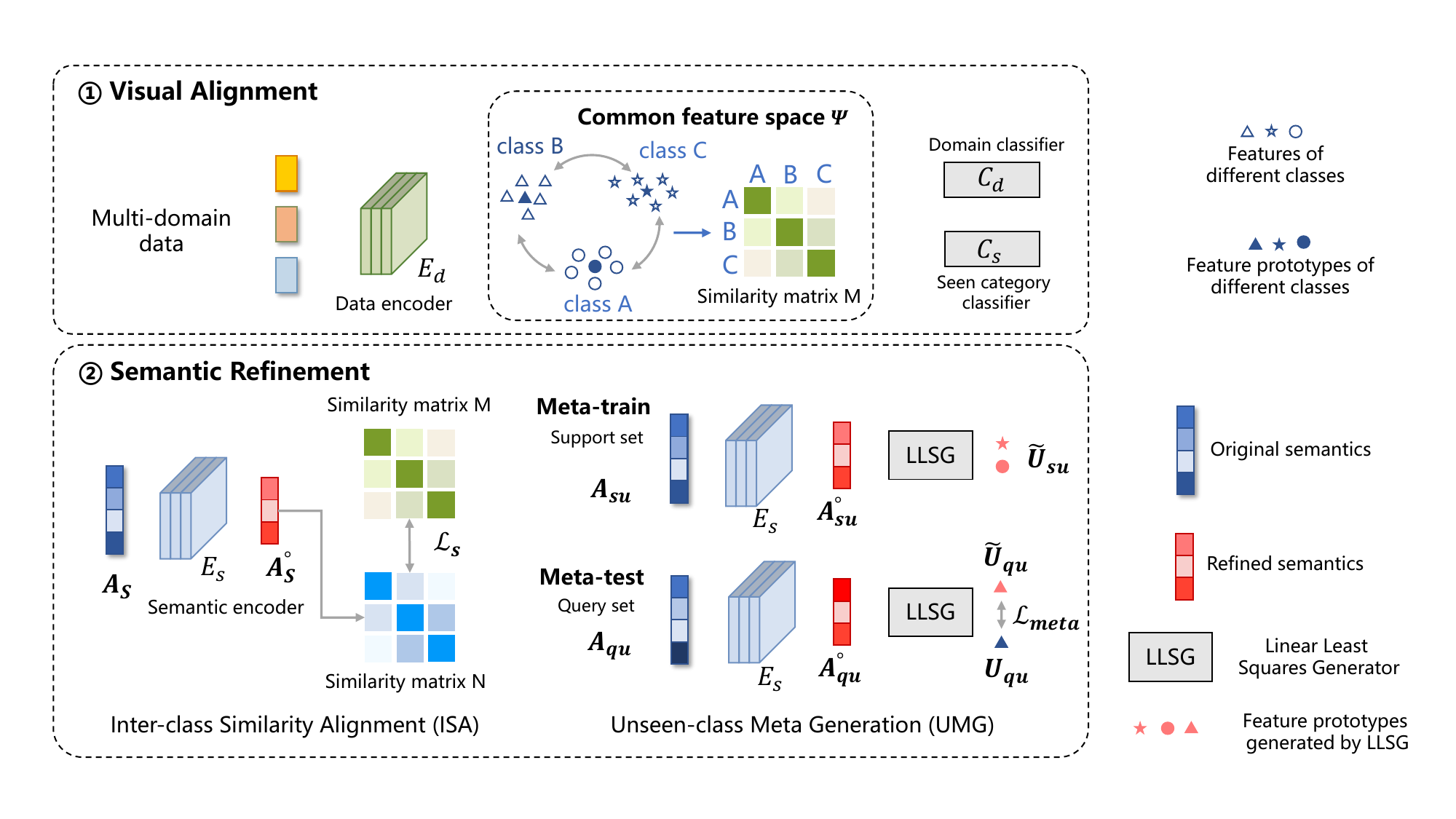}
        \caption{The framework of proposed Meta Domain Alignment Semantic Refinement (MDASR). }
        \label{fig: framework}
\end{figure*}

\section{Methodology}

In this section, we first introduce the problem formulation and notations used in this paper. From there, the process of utilizing LLMs for extracting semantics of various classes is explored. This is followed by a recapitulative explanation of the proposed MDASR. Then, the detailed paradigm for visual alignment and meta semantic refinement is presented. We end the section with introducing the CDGZSL inference procedure.

\subsection{Problem formulation and notations}

In the task of CDGZSL, our goal is to recognize both seen and unseen classes in the unseen domain. Only labeled seen class samples from multiple seen domains and their corresponding semantics are provided during the training. 

Formally, let $\mathcal{X}$ denote the input space (such as the image space), $\mathcal{A}$ denote the semantic space (such as the text or attribute space), $\mathcal{Y}$ the set of possible classes, and $\mathcal{D}$ the set of possible domains. In the training stage, a set $\mathcal{S}={(x_i, y_i, a_i, d_i)}^n_{i=1}$ is given, where $x_i \in \mathcal{X}$, $y_i \in \mathcal{Y}^s$, $a_i \in \mathcal{A}^s$, and $d_i \in \mathcal{D}^{s}$. Note that $\mathcal{Y}^s$ and $\mathcal{A}^s$ are the labels and semantic space of seen classes, respectively, and $\mathcal{Y}^s \in \mathcal{Y}$, $\mathcal{A}^s \in \mathcal{A}$. $\mathcal{D}^{s}$ is the set of seen domains, including $K$ seen domains ($K\geq 2$), and $\mathcal{D}^{s} \in \mathcal{D}$.

Given $\mathcal{S}$, we aim at learning a function mapping an image $x$ of unseen domains $\mathcal{D}^{u} \in \mathcal{D}$ to its corresponding label in a set of classes $\mathcal{Y}$. $\mathcal{D}^{s} \cap \mathcal{D}^u = \varnothing $. Following the setting of GZSL, $\mathcal{Y}=\mathcal{Y}^s \cup \mathcal{Y}^u$, where $\mathcal{Y}^u$ are the labels space of unseen classes, and $\mathcal{Y}^s \cap \mathcal{Y}^u = \varnothing $. Note that during the testing, $\mathcal{A}^u$ are provided, which is the semantic space of unseen classes.

\subsection{Extracting semantics via LLMs with prompt engineering}

In this subsection, we outline how to obtain the semantics of various classes through LLMs. We define a prompt $P$, and for a given class \( y_i \), by embedding \( y \) into $P$, we form the input to LLMs. Consequently, the output text of the LLMs can serve as the semantic representation of that class. Mathematically, this can be expressed as \( A_i = LLM(P(y_i)) \). Next, we will introduce the construction of $P$ used in this study.

We leverage a one-shot learning approach with our prompt to obtain a comprehensive semantic description from an LLM. We construct an example within the LLM input, which guides the LLM in learning the example's characteristics and generating the required semantics. Our prompt encompasses three key components: the class $y_i$ to be queried, a demonstrative class, and the description text of the demonstrative class. The structure is as follows: ``Please provide a detailed visual feature description for the class \{ \( y_i \) \}. For your reference in answering, consider the visual feature description given for the \{Eiffel Tower\} as follows: \{a reliable and rich description text about the Eiffel Tower excerpted from Wikipedia\}." In this prompt, the Eiffel Tower serves as the demonstrative class, and the model will adjust the style of the output semantics based on the descriptive text of the demonstrative class, thus making the acquisition of semantics more controllable.

\subsection{Overall framework of MDASR}

In this section, we provide a comprehensive overview of the proposed Meta Domain Alignment Semantic Refinement (MDASR), as depicted in Fig. \ref{fig: framework}. The process begins with training samples of multiple seen domains, from which we train a data encoder which aligns different domains, denoted as $E_d$, to construct a common feature space, $\Psi$. This space encapsulates the shared and distinguishable features across the seen domains, thus offering high generalizability to the unseen domain. 

To achieve CDGZSL in $\Psi$, we need to generate reliable features of unseen classes based on their semantics. As discussed in Section \ref{sec: motivation}, it is crucial to obtain the intrinsic semantics to mitigate the issues of semantic redundancy. Consequently, we introduce a meta semantic refinement framework designed to optimize the semantic encoder $E_s$, guided by features within $\Psi$. This approach preserves valuable semantic information for future feature generation. Finally, by utilizing the generated unseen features in conjunction with real seen features, we can facilitate cross-domain generalized zero-shot classification of samples from unseen domains.

\subsection{Adversarial construction of common feature space}
To construct the common feature space across different domains, we employ the Domain-Adversarial Neural Network technique \citep{ganin2016domain}. For this part, specifically refer to the upper half of Fig. \ref{fig: framework}. We establish three networks: the data encoder $E_d$, the domain classifier $C_d$, and the seen-class classifier $C_s$, parameterized by $\theta$, $\mu$, and $\sigma$ respectively. It should be noted that only training data from seen classes in the seen domains is provided.

In particular, the data encoder $E_d$ maps data from various seen domains into the feature space, subsequently engaging in a min-max game with the domain classifier $C_d$. The role of $C_d$ is to discern the domain of different features, while $E_d$ modifies the feature space to confuse $C_d$. Concurrently, the seen-class classifier $C_s$ ensures that the features remain classifiable. The loss functions of $C_d$ and $C_s$ can be represented as follows:

\begin{equation}
        \mathcal{L}_{d} = -\sum_{i=1}^k 1[y_i^d]log(C_d(E_d(x_i)))
\end{equation}

\begin{equation}
        \mathcal{L}_{c} = -\sum_{i=1}^k 1[y_i^c]log(C_s(E_d(x_i)))
\end{equation}
Here, $x_i$ denotes the $i-th$ training sample, $k$ the total number of training samples, and $y_i^d$ and $y_i^c$ the domain and class labels of $x_i$ respectively. $1[\cdot]$ serves as the indicator function.

Consequently, the total training objective of the data encoder $E_d$ can be expressed as:

\begin{equation}
        \min _{\sigma, \theta} \max _\mu \frac{1}{k}\mathcal{L}_c-\frac{\lambda}{k} \mathcal{L}_{d}
\end{equation}
where $\lambda$ is a trade-off coefficient. We designate the feature space generated by the data encoder $E_d$ as $\Psi$, housing the shared features of the seen domains.

\subsection{Meta semantic refinement in common space}

As emphasized in the motivation, if the semantic redundancy cannot be eliminated, the resulting information asymmetry would reduce the quality of the generated features for unseen classes. Therefore, in this subsection, we propose a meta semantic refined method, which includes two parts: The first part is inter-class similarity alignment, which can eliminate non-intrinsic semantics that could fail due to domain changes, but it may also inadvertently remove intrinsic information. The second part is unseen-class meta generation, which protects the intrinsic semantic information from being eliminated and maintains the ability of refined semantics to connect seen and unseen classes.

\subsubsection{Inter-class similarity alignment (ISA)}

Due to the presence of non-intrinsic semantics, the inter-class similarity obtained directly from the original semantics differs from the inter-class similarity obtained in the data-driven feature space $\Psi$. Therefore, if we can refine the semantics based on the generalizable inter-class similarity in the common space $\Psi$, we can indirectly remove the non-intrinsic semantics. Specifically, we construct the inter-class similarity leveraging the dark knowledge contained in the output of classifier $C_s$. The reason for using dark knowledge is that studies have shown that irrelevant information, like background details, still persists in the feature space \citep{cheng2020explaining}. In contrast, dark knowledge, found in the classifier's final layer, could better represent the essential information in the feature space related to class semantics because it bases the perspective of the final classification. Therefore, we can leverage the class similarity relationships calculated by dark knowledge and the similarity relationships calculated by semantics to establish the refinement constraint. The content of this part is displayed in the lower left corner of Fig. \ref{fig: framework}.

First, we come to the dark knowledge similarity part. Specifically, we obtain the dark knowledge in the form of logits. Logits are the raw, unnormalized outputs of a softmax classifier that represent the model's predicted log-odds for each class before being transformed into probabilities by the softmax function. We compute the average dark knowledge for the t-th seen class $P_t$ as:

\begin{equation}
        P_t = \frac{1}{n_t} \sum_{i=1}^{n_t} C_{s}^{\prime}(E_d(x_i^t))
\end{equation}
where $C_{s}^{\prime}$ represents the logits of $C_s$ passed through a softmax with a temperature coefficient of $T$, and $n_t$ is the number of samples for the t-th seen class. This can be considered as a kind of embedding of the class.

Subsequently, the seen-class similarity matrix $M$ in the common feature space $\Psi$ can be calculated as:

\begin{equation}
        M(i, j) = cos(P_i, P_j)
\end{equation}
$cos$ represents the cosine similarity, and $M$ represents the similarity of each class of dark knowledge prototype in $\Psi$.

Next, we come to the semantic similarity part. We build a semantic encoder $E_s$ for semantic refinement. It should be noted that our approach is applicable to different kinds of semantics. If the semantics are text extracted from LLMs, the initial $E_s$ could be a pre-trained sentence-Bert \citep{reimers2019sentence} that maps a paragraph to a fixed-length vector. If the semantics are attributes provided by human experts, $E_s$ could be a multi-layer perceptron network. With the semantic encoder $E_s$, the refined semantics can be expressed as:
\begin{equation}
        A^{\circ }=E_s(A)
\end{equation}
 Based on this, we can also obtain the similarity matrix $N$ of refined semantics $A^{\circ }$:

\begin{equation}
        N(i, j) = cos(A^{\circ }_i, A^{\circ }_j)
\end{equation}
$A^{\circ }_i$ and $A^{\circ }_j$ represent the refined semantics of the i-th seen class and the j-th seen class, respectively.

After the non-intrinsic semantics are eliminated, the inter-class similarity of semantics will become closer to the similarity in the feature space, because the common feature space does not include non-intrinsic semantic information. Therefore, if $M$ and $N$ are closely aligned, the semantic redundancy could be mitigated. The related loss function can be expressed as:

\begin{equation}
        \mathcal{L}_{s} = \Vert M-N \Vert
\end{equation}

\subsubsection{Unseen-class meta generation (UMG)}

Refining semantics solely at the level of inter-class similarity is not enough, because it could result in the loss of intrinsic information that connects seen to unseen classes. In other words, it is necessary to ensure that the intrinsic semantics are preserved and generators trained with the refined semantics can synthesize fake unseen features that are minimally different from the real ones. However, since the real unseen features are inaccessible during the training phase, guaranteeing the above condition is challenging. To address this challenge, we can simulate this process within the training set to ensure the preservation of intrinsic semantics.

In order to simulate the ZSL generation process, we implement our semantic refinement with meta-learning. The content of this part is displayed in the lower right corner of Fig. \ref{fig: framework}. Differing from traditional machine learning, the fundamental unit of meta-learning is a task, not a sample. For each task $\mathcal{T}_i=\{\mathcal{T}_{su}, \mathcal{T}_{qu}\}$, drawn from the task distribution $p(\mathcal{T})$, the support set $\mathcal{T}_{su}$ comprises $N_{su}$ seen classes, while the query set $\mathcal{T}_{qu}$ includes $N_{qu}$ seen classes. To mirror the ZSL process, the classes within the support set and the query set are disjoint. Generators that fit the features of support classes with the semantics refined by $E_s$ should be capable of generating reliable features of query classes. For each pair of support and query sets, we can generate the refined support set semantics using $E_s$, and then fit the support set and validate it with the query set for these respective tasks. Therefore, we could employ the loss to optimize semantic encoder $E_s$ to ensure it preserve intrinsic semantics beneficial to generation.

Different with exsiting meta-learning work, within each task, we utilize $E_s$ to obtain refined support set semantics $A^{\circ }_{su}$ and then train a generator from support classes. After that, fake query features are synthesized by refined query semantics and the above generator. Since real query features are at hand, loss of the query set can be calculated. We employ the loss of each task's query set to evaluate the effectiveness of $E_s$. If the loss on each task is relatively small, this indicates that the current $E_s$ is near the optimality. Otherwise, we could optimize $E_s$ using the gradient of this loss with respect to $E_s$. In order to make the loss of the query set have an effective gradient with respect to $E_s$, a linear least squares generator (LLSG) $z$ is leveraged here, where the independent variable is the refined class semantics and the dependent variable is the prototype of class features, as only a linear generator can obtain the following analytic solution:
\begin{equation}
        U_{su}=A^{\circ }_{su}z+\xi 
\end{equation}
\begin{equation}
        z^{*}=(A^{\circ }_{su})^{+}U_{su}
\end{equation}
\begin{equation}
        (A^{\circ }_{su})^{+}=((A^{\circ }_{su})^TA^{\circ }_{su}+\alpha I)^{-1}(A^{\circ }_{su})^T
\end{equation}
where $U_{su}$ is the prototypes of support class features (which are their centers), $\xi$ is the residual error, $(A^{\circ }_{su})^{+}$ is the pseudo-inverse of $A^{\circ }_{su}$, and the solution of $z$ is $z^{*}$. $\alpha$ is a small number to ensure the success of the inversion.

The fake query feature prototypes $\tilde{U}_{qu}$ produced by $z^{*}$ can be presented as: 

\begin{equation}
        \tilde{U}_{qu}=E_s(A_{qu})z^{*}
\end{equation}
Therefore, the loss of the query set can be calculated as follows:
\begin{equation}
        \label{eq: L_q}
        \mathcal{L}_q= \Vert U_{qu}-\tilde{U}_{qu} \Vert 
\end{equation}
It is easy to find (\ref{eq: L_q}) is only differential to the parameter of $E_s$, and the meta-loss can be obtained from each task and then optimizing $E_s$.
\begin{equation}
        \mathcal{L}_{meta}=\frac{1}{n_T}\sum \mathcal{L}_q
\end{equation}
where $n_T$ is the number of sampled tasks.

Considering the similarity constraint, the final loss of meta semantic refinement can be represented as:

\begin{equation}
        \mathcal{L}_{total}= \mathcal{L}_{meta} + \mathcal{L}_{s}
\end{equation}

With the optimized $E_s$, we can obtain the refined semantics $A^{\circ }$, including the seen and unseen classes, which are $A^{\circ }_s=E_s(A_s)$ and $A^{\circ }_u=E_s(A_u)$. 

\subsubsection{Feature generation with refined semantics}

At this stage, we will train a formal generator that can synthesize features in the common space under the condition $A^{\circ }$. Then, the fake unseen features and real seen features are leveraged together to obtain the final classifier. Specifically, we utilized WGAN-GP \citep{gulrajani2017improved} as our formal generator and the training process is shown as follows:
\begin{equation}
\begin{aligned}
        \mathcal{L}_{WGAN-GP} = & E[D(f)] 
        - E[D(\tilde{f})] \\
        & - \beta E[\Vert \nabla_{\hat{f}}D(\tilde{f})\Vert_2-1)^2]
\end{aligned}
\end{equation}
\begin{equation}
        \min _{G} \max _{D} \mathcal{L}_{WGAN-GP}
\end{equation}
where $E$ represents the empirical expectation, $G$ and $D$ represent the generator and discriminator of WGAN-GP, $f$ represents the real features, and $\tilde{f}$ represents the generated features, and $\beta$ is a hyperparameter. 

The generator is trained on seen class data conditioned on the refined semantics $A^{\circ}$. Because of the refined semantics, the information asymmetry between semantic space and feature space is solved beforehand, and this decreases the difficulty of the generator mapping.

After training the generator, we obtain the fake unseen features through:
\begin{equation}
        \tilde{f}_u = G(A^{\circ}_u, \epsilon)
\end{equation}
where $\epsilon$ is the randomly sampled Gaussian noise.

Combining the real seen features and the fake unseen features, the CDGZSL classifier could be trained.

\begin{equation}
        C_{c} = MLP(\tilde{f}_u, f_s)
\end{equation}
where $MLP$ is a multi-layer perceptron net.

\subsection{The CDGZSL inference procedure}

In this last subsection, we will introduce the inference process of MDASR. For testing data $x_t$, which comes from the unseen domain and could belong to seen and unseen classes, we first map it to the common feature space with data encoder $E_d$, and then predict its label with CDGZSL classifier $C_{c}$. The whole process can be formalized as follows:

\begin{equation}
        \hat{y} = C_{c}(E_d(x_t))
\end{equation}
Because $C_{c}$ is trained by both real seen features and fake unseen features with less semantic redundancy, the domain shift problem could be further alleviated, and we could gain a more accurate classifier of CDGZSL.

\section{Experiment}
In this section, we have verified the proposed MDASR on two public datasets, Office-Home\citep{venkateswara2017deep} and Mini-DomainNet\citep{9540778}. At the same time, we have also carried out a series of comparative experiments, and ablation studies to further demonstrate the effectiveness of the proposed method.

\subsection{Datasets and relative semantics}

Different from the existing CDZSL studies which utilize word vector\citep{church2017word2vec} to serve as semantics, we employ LLMs to extract semantics for the two datasets. In this way, we can validate the performance of our proposed methodology in refining LLM semantics. Due to the relatively small number of categories in Office-Home, in addition to LLM semantics, we also provide a manually annotated class-attribute matrix, which is the traditional form of semantics for ZSL, used for comparison with the semantics provided by LLM. For the convenience of other researchers, we provide all the above semantics to the public.\footnote{\url{https://github.com/chunhuiz/Semantics-of-OfficeHome-and-MiniDomainNet}}

\subsubsection{Office-Home}

The Office-Home dataset is a popular dataset used for multi-domain transfer learning tasks. It contains about 15,500 images that belong to 65 different object classes. The images are sourced from four significantly different domains: \textit{Art, Clipart, Product,} and \textit{Real-world}. As mentioned earlier, in the semantic annotations of the Office-Home dataset, we employed two forms: LLM annotation and manual annotation. For LLM annotation, as mentioned in Section III.B, we employed GPT-3.5 \citep{kocon2023chatgpt} as the LLM to obtain descriptive paragraphs as semantics for each class. For manual annotation, we first predefined 21 attributes, including the material, size, color, etc., and annotated the 65 classes, forming a $65 \times 21$ class-attribute matrix, with elements being 0 or 1. 

\subsubsection{Mini-DomainNet}

The Mini-DomainNet is a large-scale dataset which is simplified from the DomainNet \citep{peng2019moment} dataset.  It contains about 140000 images that belong to 126 different object classes. The images are sourced from four diverse domains: \textit{Art, Clipart, Sketch,} and \textit{Real}. Compared to Office-Home, Mini-DomainNet has more classes, consumes more computational resources, and poses greater challenges. Due to the large number of classes in this dataset, we did not perform manual annotations, so we only employed the LLM for semantic annotation.

\begin{table}
  \centering
  \small
  
  \caption{The CDGZSL results on Office-Home dataset}
  \begin{threeparttable}
    \renewcommand{\arraystretch}{1.3}
      \label{Tab: office}
      \begin{adjustbox}{width=\textwidth}
      
      \begin{tabular}{c|c|c|ccc|ccc|ccc}
          \hline
          \multirow{2}{*}{Semantics} & \multicolumn{2}{c|}{\multirow{2}{*}{Methods}} & \multicolumn{3}{c|}{Art} & \multicolumn{3}{c|}{Clipart} & \multicolumn{3}{c}{Product} \\
          \cline{4-12}
          & \multicolumn{2}{c|}{} & S & U & H & S & U & H & S & U & H \\
          \hline
          \multirow{8}{*}{LLM text} & \multirow{3}{*}{CuMix} & img-only & 59.43 & 0 & 0 & 50.91 & 3.28 & 6.16 & 66.21 & 0.72 & 1.42 \\
          \cline{3-12}
          & & two-level & 57.37 & 1.85 & 3.58 & 50.77 & 3.61 & 6.74 & 66.88 & 0.70 & 1.38 \\
          \cline{3-12}
          & & original & 57.83 & 2.46 & 4.71 & 51.19 & 3.94 & 7.31 & 65.09 & 1.80 & 3.50 \\
          \cline{2-12}
          & \multicolumn{2}{c|}{S-AGG} & 55.09 & 1.23 & 2.40 & 49.90 & 7.56 & 13.13 & 63.38 & 2.52 & 4.84 \\
          \cline{2-12}
          & \multicolumn{2}{c|}{SEIC} & 56.64 & 2.48 & 4.75 & 48.49 & 8.55 & 14.53 & 63.19 & 3.97 & 7.47 \\
          \cline{2-12}
          & \multicolumn{2}{c|}{ZeroDiff} & 45.20 & 27.16 & 33.93 & 37.01 & 13.16 & 19.41 & 63.73 & 19.86 & 30.28 \\
          \cline{2-12}
          & \multicolumn{2}{c|}{VADS} & 43.85 & 16.05 & 23.50 & 40.11 & 25.01 & 30.80 & 60.51 & 2.89 & 5.51 \\
          \cline{2-12}
          & \multicolumn{2}{c|}{MDASR} & 55.10 & 50.62 & \textbf{52.76} & 52.67 & 36.18 & \textbf{42.89} & 73.19 & 38.26 & \textbf{50.25} \\
          \hline
          \multirow{8}{*}{Attribute} & \multirow{3}{*}{CuMix} & img-only & 43.81 & 4.93 & 8.86 & 41.10 & 2.96 & 5.52 & 56.49 & 2.52 & 4.82 \\
          \cline{3-12}
          & & two-level & 43.45 & 5.56 & 9.85 & 39.35 & 4.60 & 8.23 & 55.17 & 6.49 & 11.61 \\
          \cline{3-12}
          & & original & 44.28 & 6.17 & 10.83 & 38.95 & 7.23 & 12.19 & 53.16 & 9.38 & 15.94 \\
          \cline{2-12}
          & \multicolumn{2}{c|}{S-AGG} & 45.31 & 5.60 & 9.96 & 37.23 & 8.89 & 14.35 & 50.41 & 7.94 & 13.71 \\
          \cline{2-12}
          & \multicolumn{2}{c|}{SEIC} & 43.50 & 6.27 & 10.96 & 41.41 & 5.26 & 9.33 & 54.70 & 6.85 & 12.17 \\
          \cline{2-12}
          & \multicolumn{2}{c|}{ZeroDiff} & 45.55 & 17.90 & 25.70 & 38.61 & 24.34 & 29.86 & 56.93 & 36.10 & 44.19 \\
          \cline{2-12}
          & \multicolumn{2}{c|}{VADS} & 35.25 & 27.78 & 31.07 & 40.67 & 18.09 & 25.04 & 61.62 & 14.08 & 22.92 \\
          \cline{2-12}
          & \multicolumn{2}{c|}{MDASR} & 59.10 & 38.27 & \textbf{46.45} & 53.28 & 30.26 & \textbf{38.60} & 72.99 & 40.07 & \textbf{51.74} \\
          \hline
      \end{tabular}
    \end{adjustbox}
      \begin{tablenotes}
          \footnotesize
          \item[*] The bold values represent the highest H-accuracy in each part.
      \end{tablenotes}
  \end{threeparttable}
\end{table}

\subsection{Experimental details}

In CDGZSL, there are both seen and unseen domains, as well as seen and unseen classes. For Office-Home, the allocation of seen and unseen domains is as follows: Art, Clipart, and Product are each designated as the unseen domain in turn. When a domain is designated as the unseen domain, the other three domains serve as the seen domains. Because we employ Resnet-50 pretrained on Imagenet\citep{deng2009imagenet} as the initial parameters for the data encoder $E_d$, only the Real-world domain is not used as the unseen domain (since it is more similar to Imagenet) among these four domains. Additionally, the division of seen and unseen classes is as follows: we select 5 classes that do not appear in Imagenet as unseen classes, 5 classes as validation, and the remaining 55 classes as seen classes. For Mini-DomainNet, similarly, aside from the Real domain, the other three domains each take a turn as the unseen domain once. Also, we select 13 classes that do not appear in Imagenet as unseen classes, 13 classes as validation, and the remaining classes as seen classes.

For the various networks used in the experiments, the specific details are as follows: The data encoder $E_d$ is the encoder of Resnet50, with initial parameters trained on Imagenet. The domain classifier $C_d$, seen class classifier $C_s$, and the final CDGZSL classifier $C_c$ are all single-layer fully connected layers followed by Softmax. The semantic encoder $E_s$ is divided into two cases: When the semantics are text generated by LLM, $E_s$ is sentence-Bert \citep{reimers2019sentence}, with initial parameters using all-MiniLM-L6-v2, and only the last two layers are fine-tuned during the training. When the semantics are manually annotated attributes, $E_s$ is a two-layer fully connected layer with an output dimension of 15. The generator and discriminator of WGAN-GP are also both two-layer fully connected layers. The activation function of all the above-mentioned fully connected layers is ReLU. For the optimizers and various hyperparameters used in training, the specific details are as follows:  \(\lambda = 1\), \(T = 5\), \(\alpha = 0.1\), \(\beta = 1\), the optimizer is Adam, and the learning rate is \(2 \times 10^{-4}\).

For the accuracy metric, we adopt three indicators: Seen class accuracy S, which is the ratio of seen class samples in the unseen domain being correctly classified; Unseen class accuracy U, which is the ratio of unseen class samples in the unseen domain being correctly classified; H-accuracy, the harmonic mean of S and U, i.e., H = 2SU/(S+U). Consistent with previous GZSL work, among these, the one that best reflects the performance of the CDGZSL algorithm is the H-accuracy.

\subsection{Results on Office-home and Mini-DomainNet}

% We have adopted six comparative methods to compare with the proposed MDASR. Specifically, these six methods can be divided into CDZSL methods (including CuMix\citep{mancini2020towards} and their two variants (img-only and two-level), S-AGG \citep{chandhok2021structured}, SEIC \citep{mondal2022seic}), and generative method modified to apply in various domains (f-CLSWGAN \citep{xian2018feature}). CDZSL methods encompass the current approaches that aim to identify unseen classes in unseen domains. However, due to the influence of the domain shift problem, they are not very effective at recognizing both seen and unseen classes simultaneously. Comparatively, generative methods can better overcome the impact of domain shift problem, but it is not designed for CDGZSL. These comparative methods utilize semantics in two different ways: When semantics exist in the form of an attribute matrix, they can be used directly; when semantics are in text form, they can first be converted into sentence vectors using sentence-Bert, and then utilized. Since our proposed method is a generative approach with meta semantic refinement, we also employed PCA\citep{raunak2019effective} as another semantic refinement method for comparison.

We have adopted six comparative methods to compare with the proposed MDASR. Specifically, these methods can be divided into two groups: CDZSL methods and recent generative methods for zero-shot learning. The CDZSL methods include CuMix\citep{mancini2020towards} (and its two variants, ``img-only" and ``two-level", which respectively refer to performing mixup only in the image space and performing mixup in both the image and feature spaces), S-AGG\citep{chandhok2021structured}, and SEIC\citep{mondal2022seic}. These methods aim to identify unseen classes in unseen domains but are significantly affected by the domain shift problem, leading to poor performance on unseen classes under CDGZSL settings. The generative methods we adopted for comparison include two recent state-of-the-art zero-shot learning methods: ZeroDiff\citep{ye2024exploring} and VADS\citep{hou2024visual}, which we adapted to CDGZSL. These generative methods have demonstrated strong capabilities in alleviating the domain shift problem. All comparative methods utilize semantics as follows: when semantics are provided in the form of attribute matrices, they are used directly; when semantics are textual, they are first transformed into semantic embeddings through sentence-Bert and subsequently utilized for prediction.

For the Office-home dataset, we utilized two types of semantic sources: one acquired from querying LLMs and the other manually annotated. The results for both types of semantics are displayed in Table \ref{Tab: office}. It is evident that the proposed MDASR outperforms other methods with both LLM-provided semantics and manually annotated semantics, with the former even surpassing the performance of the latter. We believe this is because the Office-home dataset consists of everyday items, which LLM is already capable of describing comprehensively, whereas manually provided semantics are limited by the annotators' effort and ability to provide a thorough description. However, expert intervention may still be necessary in some non-daily scenarios like industry and medicine. Observing the comparative methods, the existing four CDZSL methods, CuMix and its variants (``img-only" and ``two-level" respectively refer to performing mixup only in the image space and performing mixup in both the image and feature spaces), S-AGG, and SEIC, all have a lower accuracy for unseen classes due to domain shift problem. 

Among generative methods, ZeroDiff and VADS represent the recent state-of-the-art approaches that have shown promising performance in the zero-shot learning scenario. However, when adapted to the CDGZSL setting, their performances vary significantly across different domains and semantic types. Specifically, ZeroDiff generally achieves a relatively balanced performance across both semantics, while VADS exhibits notable variance in its ability to handle different semantics and domains. Nevertheless, both methods still show limited accuracy for unseen classes in unseen domains, indicating that directly applying these generative ZSL methods into CDGZSL tasks remains challenging. In contrast, our proposed MDASR consistently surpasses ZeroDiff and VADS by a considerable margin, demonstrating the effectiveness of the proposed meta semantic refinement strategy in handling semantic redundancy and domain shift simultaneously.

% It can be observed that the modified f-CLSWGAN is the current state-of-the-art. Meanwhile, the PCA semantic refinement used for comparison has negative effects on generative methods most of time. The performance can often tend towards degradation. This is because the optimization of PCA does not consider the effect of semantic redundancy, and does not take measurement to preserve effective semantics. The proposed MDASR, in contrast, enhances performance over f-CLSWGAN across different semantics and domains. 

\begin{table}[htbp]
  \centering
  \small
  \caption{The CDGZSL results on Mini-DomainNet dataset} 
  \begin{threeparttable}
    \renewcommand{\arraystretch}{1.3}
    \label{Tab: minidomainnet}
    \begin{adjustbox}{width=\textwidth}
    \begin{tabular}{c|c|ccc|ccc|ccc}
      \hline
      \multicolumn{2}{c|}{\multirow{2}{*}{Methods}} & \multicolumn{3}{c|}{Clipart} & \multicolumn{3}{c|}{Painting} & \multicolumn{3}{c}{Sketch} \\
      \cline{3-11}
      \multicolumn{2}{c|}{} & S & U & H & S & U & H & S & U & H \\
      \hline
      \multirow{3}{*}{CuMix} & img-only & 63.98 & 2.67 & 5.12 & 63.69 & 2.64 & 5.06 & 59.61 & 1.86 & 3.60 \\
      & two-level & 64.54 & 3.32 & 6.31 & 64.20 & 2.58 & 4.96 & 59.75 & 3.42 & 6.46 \\
      & original & 64.20 & 2.49 & 4.79 & 64.27 & 3.10 & 5.91 & 54.70 & 5.07 & 9.27 \\
      \hline
      \multicolumn{2}{c|}{S-AGG} & 63.98 & 2.89 & 5.53 & 62.65 & 0.90 & 1.77 & 56.96 & 1.91 & 3.69 \\
      \hline
      \multicolumn{2}{c|}{SEIC} & 64.24 & 2.01 & 3.89 & 63.46 & 1.09 & 2.14 & 58.45 & 1.08 & 2.12 \\
      \hline
      \multicolumn{2}{c|}{ZeroDiff} & 62.32 & 3.87 & 7.29 & 60.61 & 5.01 &9.26  & 57.27 & 5.07 & 9.31 \\
      \hline
      \multicolumn{2}{c|}{VADS} & 63.52 & 4.65 & 8.66 & 56.28 & 6.21 & 11.19 & 56.59 & 3.94 & 7.37 \\
      \hline
      \multicolumn{2}{c|}{MDASR} & 63.69 & 29.43 & \textbf{40.25} & 61.02 & 26.16 & \textbf{36.62} & 55.16 & 28.25 & \textbf{37.36} \\
      \hline
    \end{tabular}
  \end{adjustbox}
  \begin{tablenotes}
    \footnotesize
    \item[*] The bold values represent the highest H-accuracy in each part.
\end{tablenotes}
  \end{threeparttable}
\end{table}

\begin{figure*}[htbp]
        \centering
        \includegraphics[width=\textwidth]{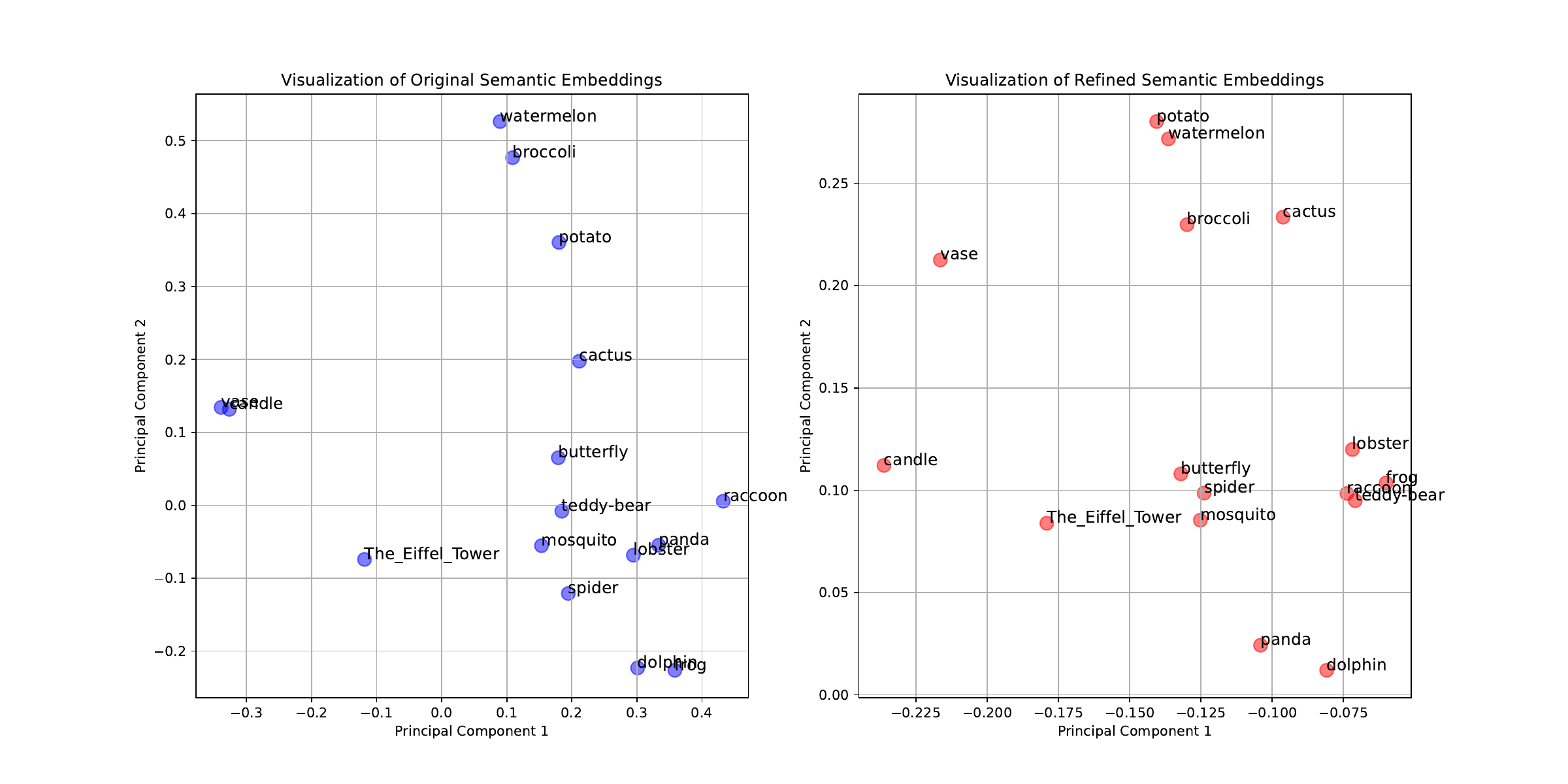}
        \caption{Visual comparison between the original semantic embeddings and the semantic embeddings refined by MDASR.}
        \label{fig: PCA_semantic}
\end{figure*}

For the Mini-DomainNet dataset, due to the larger number of classes, we only demonstrate the results using semantics automatically acquired from LLMs, which are presented in Table \ref{Tab: minidomainnet}. Given that Mini-DomainNet has a greater number of unseen classes, the difficulty of CDGZSL is also higher. The proposed MDASR achieves an H-accuracy ranging from 36.62\% to 40.25\% in different unseen domains, which remains the highest among all the methods compared. The results of the other methods are similar to those on Office-Home but with a general decline in H-accuracy. The domain shift problem effect is still severe in CDZSL methods. Under the CDZSL setting, their accuracy for unseen classes may reach 8.50\%-25.90\% \citep{mondal2022seic}, but under the CDGZSL setting, it is only about 2\%-5\%. Among the generative zero-shot learning methods, ZeroDiff and VADS exhibit moderate performance in alleviating the domain shift problem on Mini-DomainNet. However, due to the complexity and large category number of Mini-DomainNet, the improvement brought by these methods is still limited under the CDGZSL setting. In contrast, the proposed MDASR significantly outperforms ZeroDiff and VADS, showcasing strong generalization across various domains and semantics. This result further validates the robustness and effectiveness of our meta semantic refinement method in learning intrinsic semantic representations.

% f-CLSWGAN still maintains a certain capacity for alleviating domain shift problem. If our MDASR is leveraged, the H-accuracy is further improved compared to f-CLSWGAN; however, if PCA is used for semantic refinement, since its objective function is to maximize variance, which is unrelated to the intrinsic semantics, it usually brings about a negative effect. Overall, in the two datasets, our proposed MDASR achieved the highest H-accuracy, demonstrating the capacity to recognize seen and unseen classes in unseen domains. 

Additionally, to further illustrate the effectiveness of the proposed MDASR in eliminating redundant semantics, we performed a two-dimensional visualization of the original semantic embeddings and the refined semantic embeddings, as shown in Fig. \ref{fig: PCA_semantic}. Since the Mini-domainnet category count exceeds 100, considering spatial constraints and presentation effectiveness, we randomly selected 17 categories for display. From the figure, it can be seen that the refined semantics are more closely aligned with intrinsic visual features rather than other redundant information. For example, initially, in the original semantics, watermelon and broccoli are very close to each other but far from potato, which may suggest a large proportion of color in the original semantics since the first two are green. However, after refinement, the distance between watermelon and potato has decreased, indicating that the new semantics are no longer influenced by color, which can easily change with domain, but instead refer to `shape' that is more diffcult to alter as both are elliptically shaped. Furthermore, in the original semantics, vase and candle nearly overlap, which is inconsistent with visual features; the refined semantics also separate the two. Lastly, in the original semantics, butterfly, mosquito, and spider are far apart, but the refined semantics bring them closer to each other, which is also consistent with visual features. Overall, our method is effective in eliminating redundant components in semantics under the supervision of the common feature space.

\subsection{Ablation Study}

\begin{figure*}[htbp]
  \centering
  \includegraphics[width=\textwidth]{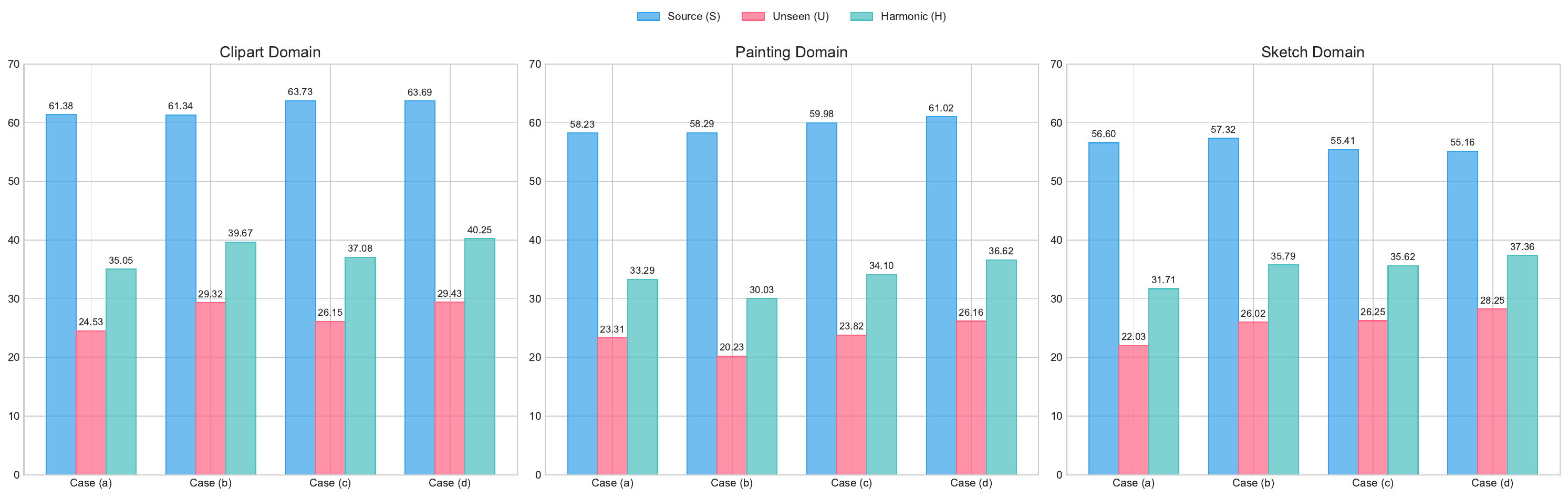}
  \caption{The ablation experiment results on Mini-DomainNet dataset}
  \label{fig: ablation}
\end{figure*}

We conducted ablation experiments on the Mini-DomainNet dataset to prove the effectiveness of each component in our proposed method. The relevant results are presented in Fig. \ref{fig: ablation}. We conducted a total of four groups of experiments: Case (a): without ISA and UMG, Case (b): with ISA, without UMG, Case (c): without ISA, with UMG, Case (d): with both ISA and UMG. Specifically, case (a) represents a generative method without any semantic refinement in the aligned space, and it results in 33.35\% accuracy on average. Next, we verified the two parts of the proposed meta semantic refinement loss functions: the similarity matrix alignment $\mathcal{L}_{s}$ and the meta-learning loss $\mathcal{L}_{meta}$. It can be observed that using only the similarity matrix alignment, as in case (b), some results improve while others worsen, suggesting that this method might lose certain important semantics. Solely utilizing meta-learning does not lead to degradation but offers limited improvement, as shown in case (c). Only by using both, as in case (d), do we achieve the best semantic refinement results, with the average accuracy of 38.07\%.

\section{Conclusion}

In this paper, we have introduced the task of cross-domain generalized zero-shot learning towards unseen domains for the first time and developed the MDASR approach which addresses the information asymmetry caused by the redundancy of LLM-based semantics. We construct a common feature space of seen domains, enabling the discriminative relations learned from the seen domains to generalize to unseen domains. Under the guidance of the common feature space, by designing ISA which aligns intra-class relationship, we have achieved the removal of non-intrinsic semantics. Meanwhile, through UMG that simulates the synthesis of unseen features, we retain intrinsic semantics that could connect the seen and unseen classes. Building on this, our proposed method can generate features of unseen classes by leveraging intrinsic semantics shared across domains, alleviating domain shift problem and recognizing both seen and unseen classes in unseen domains. We have validated the CDGZSL capabilities of the proposed method on two open-source datasets. On the Office-Home dataset, the average H-accuracy of CDGZSL reached 47.11\%, and on the more challenging Mini-DomainNet dataset, the average H-accuracy of CDGZSL reached 38.07\%. We have also made the corresponding semantics available to the public, providing a benchmark for future studies.

%% The Appendices part is started with the command \appendix;
%% appendix sections are then done as normal sections
%% \appendix

%% \section{}
%% \label{}

%% If you have bibdatabase file and want bibtex to generate the
%% bibitems, please use
%%
%%  \bibliographystyle{elsarticle-num} 
%%  \bibliography{<your bibdatabase>}

%% else use the following coding to input the bibitems directly in the
%% TeX file.

% \section*{CRediT authorship contribution statement}
% Liangjun Feng: Writing - original draft, Experiment- Data processing, Figure plotting. Jiancheng Zhao: Writing-review and editing. Jiaqi Yue: Writing-Polishing the English presentation. Chunhui Zhao: Methodology-Proponents of major academic ideas and supervision.

\section*{Declaration of competing interest}
The authors declare that they have no known competing financial interests or personal relationships that could have appeared to
influence the work reported in this paper.

\section*{Acknowledgments}
This work was supported in part by the National Natural Science Foundation of China (No. 62450020), and in part by the National Science Fund for Distinguished Young Scholars (No. 62125306), and in part by Zhejiang Key Research and Development Project (2024C01163), and in part by the Open Research Project of the State Key Laboratory of Industrial Control Technology, China (Grant No. ICT2024B19).

\bibliographystyle{elsarticle-harv} 

\bibliography{sn-bibliography}

\end{document}